\documentclass{article}

\PassOptionsToPackage{numbers, compress}{natbib}
\usepackage[final]{style}

\usepackage{graphicx}
\usepackage{subcaption}
\usepackage[utf8]{inputenc} % allow utf-8 input
\usepackage[T1]{fontenc}    % use 8-bit T1 fonts
\usepackage{hyperref}       % hyperlinks
\usepackage{url}            % simple URL typesetting
\usepackage{booktabs}       % professional-quality tables
\usepackage{amsfonts}       % blackboard math symbols
\usepackage{nicefrac}       % compact symbols for 1/2, etc.
\usepackage{microtype}      % microtypography
\usepackage{xcolor}         % colors

\title{Internet of Things Meets Robotics: A Survey of Cloud-based Robots}

\author{%
  Chrisantus Eze \\
  Department of Computer Science\\
  Oklahoma State University\\
  Stillwater, Oklahoma \\
  \texttt{chrisantus.eze@okstate.edu} \\
}

\begin{document}

\maketitle

\begin{abstract}
  This work presents a survey of existing literature on the fusion of the Internet of Things (IoT) with robotics and explores the integration of these technologies for the development of the Internet of Robotics Things (IoRT). The survey focuses on the applications of IoRT in healthcare and agriculture, while also addressing key concerns regarding the adoption of IoT and robotics. Additionally, an online survey was conducted to examine how companies utilize IoT technology in their organizations. The findings highlight the benefits of IoT in improving customer experience, reducing costs, and accelerating product development. However, concerns regarding unauthorized access, data breaches, and privacy need to be addressed for successful IoT deployment.
\end{abstract}

\section{Introduction}
\label{sec:introduction}
The Internet of Things has continued to experience increased adoption, with numerous research efforts focusing on its deployment in various aspects of our daily lives. Similarly, robotics, which has been around for a while, has played a crucial role in diverse domains. However, there was a period during which both fields underwent significant development independently. It has now become evident that current scenarios require the integration of these two disciplines and a collaborative approach from their respective communities.

Over the years, areas such as healthcare and agriculture have witnessed significant impacts from IoT and robotics. \cite{Pradhan2021InternetOT} discussed how IoT and robotics have transformed healthcare services, including rehabilitation, assistive surgery, elderly care, and prosthetics. In another study by \cite{10119284}, a model called Automatic Agricultural field Robot – Agro-bot was introduced, which utilizes robotics and automation to perform various agricultural operations such as soil digging, seed sowing, precise watering, spraying, and weeding.

\begin{figure*}[htbp] % htbp stand for "here", "top", "bottom", "page"
\includegraphics[scale=0.50]{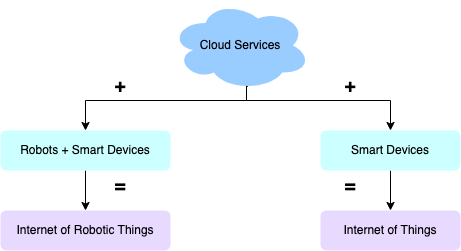} 
\caption{This illustrates the relationship between IoT and IoRT}
\label{fig:iort}
\end{figure*}

This paper provides an overview of IoT and Robotics technologies and explores their integration for the development of the Internet of Robotics Things (IoRT). The convergence of IoT and robotics is evident in various distributed and heterogeneous robot control paradigms, such as network robot systems \cite{sanfeliu2008network} and robot ecologies \cite{4650962}. Additionally, approaches like ubiquitous robotics \cite{4209522} and cloud robotics \cite{7006734, 6201212, 6201213} have emerged, utilizing server-side resources to handle resource-intensive functionalities. The concept of the "Internet of Robotic Things" (IoRT) was introduced to describe a scenario where sensor data from diverse sources is fused and processed using local and distributed intelligence to control physical objects. This cyber-physical perspective of IoRT leverages IoT's sensor and data analytics technologies to enhance robots' situational awareness, resulting in improved task execution. Notable use cases include intelligent transportation  \cite{roy2017iot} and companion robots \cite{simoens2016internet}. Subsequent literature on IoRT has explored alternative perspectives, such as emphasizing robust team communication \cite{razafimandimby2016neural} and considering robots as additional sensors within the IoT ecosystem \cite{7992744, Scilimati2017IndustrialIO}.

Furthermore, cloud computing now finds increased applications in human-robot interaction, which is a field that seeks to develop techniques to enhance how humans interact and collaborate with robots. Notable works in this area include those by \cite{6129046, podpora2020human, 8279345}. \cite{6129046} proposed a system that provides companionship to the elderly using robotics, which can be controlled physically or remotely via cloud infrastructures. \cite{podpora2020human} presents a detailed concept of Human-Robot Interaction systems architecture, focusing on acquiring information about the robot's interlocutor. It includes an IoT-based sensor subsystem for interlocutor identification and integrates additional subsystems for human and visual identification. \cite{8279345} proposes a new system architecture for SIGVerse, addressing key problems in robotics simulation. By adopting Unity middleware for VR applications and ROS for robot software development, the proposed architecture enables faster development of human-robot interaction applications. Additionally, \cite{eze2023enhancing} proposed techniques and guidelines to design Augmented Reality (AR) interfaces, enhancing the experience of humans interacting with robots.

The rest of the paper is organized according to the following: the existing literature is reviewed in Section \ref{sec:literature}, Section \ref{sec:challenges} presents some of the core challenges being faced, and finally, in Section \ref{sec:conclusion} the work is summarized and concluded.

\section{Review of Literature}
\label{sec:literature}
In this section, I will explore some existing works that focus on integrating IoT with robotics for various applications in healthcare, agriculture, and other fields.

\subsection{Design and Implementation of Ploughing and Seeding of Agriculture Robot Using IOT}
\cite{Poonguzhali2020DesignAI} proposed a system that involves the utilization of a robot for automating the processes of sowing crops, plowing the field, and automatically sprinkling water on the crops. The robot is equipped with motors that enable its movement in both forward and backward directions. To train the robot, an embedded instructing coding system is employed, enabling it to carry out the tasks of sowing, plowing, and watering the crops. By setting specific time intervals, these operations can be executed automatically without the need for manual labor. 

\subsection{Internet of Things-Based Devices/Robots in Agriculture 4.0}
\cite{singh2022internet} examines the work conducted by researchers and engineers, as well as the major application areas of Internet of Things (IoT)-based devices in smart agriculture. The article also discusses the various equipment and technologies employed in smart agriculture. Additionally, it highlights the challenges encountered when implementing IoT-based devices in agriculture. The study's findings are valuable and helpful for researchers, agriculturists, and professionals working in the field of modern agriculture. The utilization of IoT brings benefits to farming and agriculture by introducing concepts such as live tracking, pest control, irrigation management, soil analysis, and more. Consequently, this paper provides a critical review of IoT-based devices deployed in agricultural fields, with the introduction section exploring the growth of IoT-based devices in agriculture.

\subsection{Agrobot - An IoT-Based Automated Multi-Functional Robot}
The authors in \cite{10119284} presented an Agrobot, a robot designed to meet agricultural requirements and enhance various aspects of farming operations. This Agrobot incorporates an automated seed-planting machine, thereby significantly improving agricultural output. With the ability to perform tasks such as drilling, fertilizing, seed sowing, and watering, this single robot serves as an optimal solution to overcome challenges arising from the limited availability of working capital and increasing labor costs. By introducing automated equipment in farming, it becomes possible to scale up production within a limited timeframe.

\subsection{Internet of Things and Robotics in Transforming Current-Day Healthcare Services}
The study in \cite{Pradhan2021InternetOT} examines the role of the Internet of Things (IoT) and robotics in transforming healthcare services. It presents the various functionalities of ideal IoT-aided robotic systems and highlights their importance in healthcare applications. The research focuses on the application of IoT and robotics in healthcare services such as rehabilitation, assistive surgery, elderly care, and prosthetics. It provides a detailed analysis of recent developments, the current status, limitations, and challenges in this field. Additionally, the study discusses the role and applications of these technologies in managing the COVID-19 pandemic. It offers comprehensive knowledge of the functionality, application, challenges, and future scope of IoT-aided robotic systems in healthcare services. This research provides valuable insights for future researchers aiming to enhance healthcare services using these technologies.

\subsection{Internet of Robotic Things: Context-Aware and Personalized Interventions of Assistive Social Robots}
In our daily lives, versatile and capable assistive service and companion robots are present, operating within our living environment. These robots possess the ability to manipulate physical objects, navigate their surroundings, and engage in conversations. However, human behavior is often unpredictable and dynamic, necessitating the assistance of a cloud-backend system. This system serves several purposes, including analyzing data from sensors and wearables, determining the required robotic tasks, and providing the necessary support for executing these tasks effectively in our everyday environment. The authors in \cite{simoens2016internet} presented a system architecture design for an Internet-of-Robotic-Things (IoRT) framework. To showcase the practical application of this framework, they focused on a case study involving personal interactions facilitated by a companion robot. The objective of the case study was to alleviate behavioral disturbances experienced by individuals with dementia.

\subsection{Innovative and efficient method of robotics for helping the Parkinson's disease patient using IoT in big data analytics}
Big data has accumulated a massive amount of stored data in various fields, including robotics, the Internet of Things (IoT), and healthcare systems. While IoT-based healthcare systems play a crucial role in the big data industry, there are instances where accurately predicting results through sensing can be challenging. The proposed system in \cite{sivaparthipan2020innovative}, which combines artificial intelligence and IoT for Parkinson's disease, has the potential to significantly enhance gait performance. This research provides a clear understanding of the role of robots in Parkinson's disease and their interaction with big data analytics. The research scheme involves collecting data from big data sources and introducing a novel approach that utilizes laser scanning combined with piecewise linear Gaussian dynamic time warp machine learning. A laser scanning system is used to scan the path for obstacles and identify safe areas. The primary role of the robot is to predict the motion of the patient using the walker and provide appropriate physical training. As both the patient and the robot have fixed sensors, the robot walks alongside the patient to accurately predict the patient's walker motion.

\subsection{Healthcare Robots Enabled with IoT and Artificial Intelligence for Elderly Patients}
The authors in \cite{porkodi2021healthcare} identified the needs of elderly patients and proposed solutions using personalized robots. The utilization of IoT devices enables the quick prediction of emergency situations by providing vital information, while artificial intelligence aids in suggesting necessary actions. Health data from patients is obtained through IoT-based wearable devices and processed using AI for decision-making. The designed robot then carries out the required actions based on these decisions. Humanoid robots can be designed to provide healthcare and physical assistance to elderly patients and those with chronic conditions. Additionally, animal-like robots can be developed to act as pets and offer companionship to individuals with psychosocial issues. The main objective is to review the capabilities of robots and lay the groundwork for future advancements, envisioning a robot that can prevent interventions, perform multiple functions, engage in motivational interactions, provide enhanced educational data, and promptly alert an ambulance in case of emergencies.

\subsection{Distributed Perception by Collaborative Robots}
The authors of the study in \cite{8411096} proposed a framework to leverage the combined computational power of multiple low-power robots in order to enable efficient, dynamic, and real-time recognition. The method is designed to adapt to the availability of computing devices during runtime and adjust to the inherent dynamics of the network. The framework is applicable to any distributed robot system. To demonstrate its effectiveness, the researchers implemented the framework using several Raspberry-Pi3-based robots equipped with cameras (up to 12 robots). They successfully implemented a state-of-the-art action recognition model for videos and two recognition models for images. The results show that this approach allows a group of low-power robots to achieve similar performance (measured in terms of the number of images or video frames processed per second) compared to a high-end embedded platform, Nvidia Tegra TX2.

\subsection{Robot Cloud: Bridging the power of robotics and cloud computing}
Cloud computing has become a transformative force in the cyber world, serving as a prominent computing and service platform for resource sharing. This includes sharing platforms, software applications, and various services. The authors in \cite{Du2017RobotCB} tried to merge the cyber world with the physical world through the concept of the "Robot Cloud," which aims to combine the capabilities of robotics with cloud computing. To realize this vision, the researchers propose a novel Robot Cloud stack, designed to support and facilitate the integration. They employ a service-oriented architecture (SOA) to enhance the flexibility, extensibility, and reusability of the functional modules within the Robot Cloud. To showcase their design approach, a prototype of the Robot Cloud is developed using the widely-used Google App Engine. Additionally, simulation experiments are conducted in the context of a "robot show" application scenario. These experiments evaluate the proposed scheduling policy and investigate the impact of different request distributions and robot center solutions.

\subsection{Internet of Things (IoT) based Robotic Arm}
In this project \cite{gawli2017internet}, the authors achieved control over a robotic arm not only through traditional wired controls but also by utilizing the emerging technology of the Internet of Things (IoT). By incorporating an IoT interface, they were able to remotely control the robotic arm, making it applicable to various industrial settings where machines require control from distant locations. This project not only enables real-time response to commands but also records and reproduces specific movements, thereby reducing the need for human intervention and effort in repetitive tasks.

\subsection{ROSLink: Bridging ROS with the Internet-of-Things for Cloud Robotics}
The Internet of Things (IoT) and cloud robotics may be seamlessly integrated with ROS-enabled robots thanks to a new communication protocol called ROSLink \cite{koubaa2017roslink}. ROSLink does away with the necessity for Network Address Translation (NAT) and enables any robot to be mapped to any user via the Internet by implementing the server in a widely accessible cloud. The JSON serialization-based protocol has been shown to be effective and dependable for operating robots over the cloud. Given the accessibility of fast internet connections and an abundance of bandwidth, ROSLink offers a compact and scalable solution with little extra network overhead. The evaluation research looked at how various network topologies affected performance.

\section{Challenges and Issues}
\label{sec:challenges}
Several issues could arise when smart objects and devices are interconnected with robots. This section describes some of the key issues and challenges currently facing the adoption of IoT and robotics.

\subsection{Security and Privacy}
It is important to ensure that the connection is secure and user data is protected against malicious hackers. To achieve optimal security and privacy, there is a need to explore a novel access control mechanism that works in conjunction with robot authentication, with a focus on defining and managing robot identity \cite{grieco2014iot}. Additionally, implementing algorithms for data confidentiality and message integrity is essential, along with advanced approaches to identify and restrict the involvement of untrusted devices and robots within the system.

\subsection{Computational Complexity}
Interconnecting multiple devices in a shared network often leads to various computational issues, which are further exacerbated when highly computationally intensive devices like robots are added to the network. These issues can manifest in different forms, such as implementing network protocols, designing algorithms for message sharing among devices, real-time knowledge sharing, handling big data, and optimizing bandwidth usage.

\subsection{Ethical}
There have been several ethical concerns regarding the deployment of IoT and robots in various settings. \cite{Allhoff2018TheIO} discussed ethical issues related to the deployment of IoT technology, including informed consent, privacy, trust, and physical safety. \cite{Sharkey2010GrannyAT} outlined ethical concerns associated with deploying robot applications for assisting the elderly, such as reduced human contact, feelings of objectification and loss of control, loss of privacy, loss of personal liberty, deception, and infantilization, and the circumstances of elderly people controlling robots.

\section{Summary and Conclusion}
\label{sec:conclusion}
In this study, I conducted a literature review on the integration of the Internet of Things (IoT) with robotics. The aim of this paper is to present an overview of IoT and robotics technologies and investigate their integration for the development of the Internet of Robotics Things (IoRT). The survey primarily focused on the application of IoRT in healthcare and agriculture. Additionally, I addressed some of the major concerns related to the adoption of IoT and robotics.

Furthermore, I conducted an online survey to examine the usage of IoT technology in companies. Here are the findings from the survey:

\textbf{Customer experience:} The company used IoT to provide customers with real-time information about products and services, such as product availability, and delivery status. This helps to improve customer satisfaction and loyalty.

\textbf{Costs:} IoT can be used to automate tasks, such as monitoring and controlling equipment, which can help to reduce costs. The company uses IoT to monitor and regulate the temperature in the building. They also ensure the proper regulation of the temperature in unoccupied rooms to conserve energy and cost electricity cost.

\textbf{Speed of product development:} IoT can also be used to collect data from products in the field, which can be used to improve the design and development of new products. For example, the company used IoT to collect information about how users use their products and use those data to improve the next iteration of the products. 

\hfill \break
They also stated some of the concerns they have with the deployment of IoT, such as:

\textbf{Unauthorized access:} IoT devices are often connected to the internet, which makes them vulnerable to unauthorized access.

\textbf{Data breaches:} IoT devices can collect a lot of data about users, which can be used for malicious purposes if it is not properly secured.

\textbf{Privacy concerns:} Some users may be concerned about the amount of data that IoT devices collect about them.

\hfill \break
Therefore, it is noteworthy that organizations that successfully adopt IoT need to address these challenges in order to protect their users and their data.

{\small
\bibliographystyle{unsrt}
\bibliography{iort}
}
%%%%%%%%%%%%%%%%%%%%%%%%%%%%%%%%%%%%%%%%%%%%%%%%%%%%%%%%%%%%

\end{document}